%% file: main.tex
\documentclass[runningheads]{llncs}

\usepackage[mobile]{eccv}

\usepackage{eccvabbrv}
\usepackage{graphicx}
\usepackage[accsupp]{axessibility}  % Improves PDF readability for those with disabilities.
\usepackage{capt-of,xcolor,xspace,enumitem}
\usepackage{amsmath,amssymb,amsbsy,amsfonts,dsfont,pifont,bm,bbm,mathrsfs,mathtools,nicefrac}
\usepackage{algorithm,algpseudocode,listings}
\usepackage{booktabs,multirow,adjustbox,diagbox,threeparttable,tabularray}

\usepackage[pagebackref,breaklinks,colorlinks,citecolor=eccvblue]{hyperref}

\usepackage{orcidlink}
\usepackage{wrapfig}
\usepackage[capitalize]{cleveref}  % Should be loaded after 'hyperref', and works perfectly with 'subfigure'.
\crefname{section}{Sec.}{Secs.}
\Crefname{section}{Section}{Sections}
\crefname{table}{Tab.}{Tabs.}
\Crefname{table}{Table}{Tables}
\crefname{figure}{Fig.}{Figs.}
\Crefname{figure}{Figure}{Figures}
\crefname{equation}{Eq.}{Eqs.}
\Crefname{equation}{Equation}{Equations}
\newcommand{\tocite}[1]{{\color{red} [TO CITE]}}
\newcommand{\method}{{\texttt{InFusion}}\xspace}

\begin{document}

\title{InFusion: Inpainting 3D Gaussians via \texorpdfstring{\\}~Learning Depth Completion from Diffusion Prior}
\titlerunning{\method}

\author{
    Zhiheng Liu \inst{1,4} \textsuperscript{*} \and
     Hao Ouyang\inst{2,3} \textsuperscript{*} \and
    Qiuyu Wang \inst{3} \and
    Ka Leong Cheng \inst{2,3} \and
    Jie Xiao \inst{1,4} \and
     Kai Zhu \inst{4} \and
     Nan Xue \inst{3} \and
     Yu Liu  \inst{1}\and 
     Yujun Shen  \inst{3}\and
     Yang Cao  \inst{1} \textsuperscript{†}
}
\authorrunning{Liu.~ et al.}

\institute{
    University of Science and Technology of China \and The Hong Kong University of Science and Technology \and Ant Group \and Alibaba Group
}

\let\oldthefootnote\thefootnote
\renewcommand{\thefootnote}{}
\footnotetext{*These authors contributed equally to this work. \textsuperscript{†}Corresponding author.}
% \maketitle
\renewcommand\twocolumn[1][]{#1}
\maketitle

\input{sections/0.abs.tex}

\input{sections/1.intro.tex}
\input{sections/2.related.tex}
\input{sections/3.method.tex}
\input{sections/4.exp.tex}
\input{sections/5.conclusion.tex}

\input{sections/6.ref.tex}

\input{supp_1.tex}

% \input{sections/7.appendix.tex}

\end{document}

%% file: sections/0.abs.tex
\begin{abstract}

3D Gaussians have recently emerged as an efficient representation for novel view synthesis.
This work studies its editability with a particular focus on the inpainting task, which aims to supplement an incomplete set of 3D Gaussians with additional points for visually harmonious rendering.
Compared to 2D inpainting, the crux of inpainting 3D Gaussians is to figure out the rendering-relevant properties of the introduced points, whose optimization largely benefits from their initial 3D positions.
To this end, we propose to guide the point initialization with an image-conditioned depth completion model, which learns to directly restore the depth map based on the observed image.
Such a design allows our model to fill in depth values at an aligned scale with the original depth, and also to harness strong generalizability from large-scale diffusion prior.
Thanks to the more accurate depth completion, our approach, dubbed \method, surpasses existing alternatives with sufficiently better fidelity and efficiency (\textit{i.e.}, $\sim20\times$ faster) under various complex scenarios.
We further demonstrate the effectiveness of \method with several practical applications, such as inpainting with user-specific texture or with novel object insertion.
Our code is public available at \url{https://johanan528.github.io/Infusion/}.


\end{abstract}

%% file: sections/1.intro.tex
\section{Introduction}
% Application
Recent developments in 3D representation~\cite{barron2021mip,wang2021neus,mildenhall2021nerf,kerbl20233d } have highlighted 3D Gaussians~\cite{kerbl20233d,tang2023dreamgaussian,wu20234d,chen2023text,yi2023gaussiandreamer} as an essential approach for novel view synthesis, owing to the ability to produce photorealistic images with impressive rendering speed. 3D Gaussians offer explicit representation and the capability for real-time processing, which significantly enhances the practicality of editing 3D scenes. The study of how to editing 3D Gaussians is becoming increasingly vital, particularly for interactive downstream applications such as virtual and augmented reality (VR/AR). Our research focuses on the inpainting tasks that are crucial for the seamless integration of edited elements, effectively filling in missing parts and serving as a foundational operation for further manipulations.

\begin{center}
    % \vspace{-5pt}
    \includegraphics[width=0.94\linewidth]{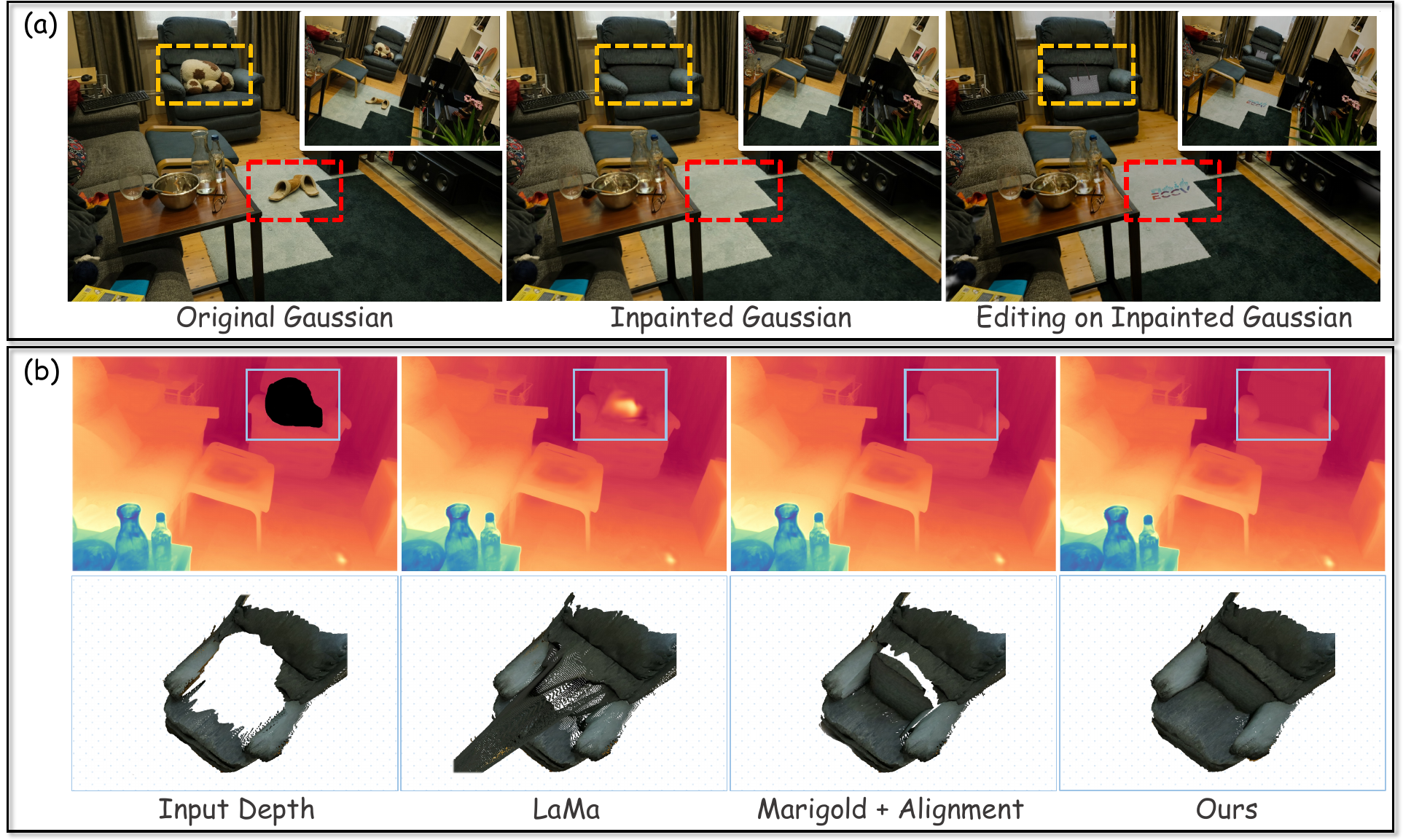}
    % \vspace{-17pt}
    \captionsetup{type=figure}
    \caption{% 
        We present \method, an innovative approach that delivers efficient, photorealistic inpainting for 3D scenes with 3D Gaussians. As demonstrated in (a), \method enables the seamless removal of 3D objects, along with user-friendly texture editing and object insertion. Illustrated in (b), \method learns depth completion with diffusion prior, significantly enhancing the depth inpainting quality for general objects. We show the visualizations of the unprojected points, which exhibit substantial improvements over baseline models~\cite{suvorov2022lama,ke2023repurposing}.
    }
    \label{fig:teaser}
    \vspace{10pt}
\end{center}

% Current Issues 
Initial explorations into 3D Gaussian inpainting have focused on growing Gaussians from the boundary of the uninpainted regions, using inpainted 2D multiview images for guidance~\cite{ye2023gaussian,chen2023gaussianeditor,fang2023gaussianeditor}. This method, however, tends to produce blurred textures due to inconsistencies in the generation process, and the growing can be quite slow. Notably, the training quality for Gaussian models is significantly improved when the initial points are precisely positioned within the 3D scene, particularly on object surfaces. A practical solution to improve the fine-tuning of Gaussians is to predetermine these initial points where inpainting will occur, thereby simplifying the overall training process. In allocating initial points for Gaussian inpainting, the role of depth map inpainting can be pivotal. The ability to convert inpainted depth maps into point clouds facilitates a seamless transition to 3D space, while also leveraging the potential to train on expansive datasets~\cite{mayer2016large,menze2018object,schuhmann2022laion}.

% Our method .\method 
To this end, we introduce \method, an innovative approach to 3D Gaussian inpainting that leverages depth completion learned from diffusion models~\cite{ avrahami2022blended,blattmann2023align,podell2023sdxl,rombach2022high}. Our method demonstrates that with a robustly learned depth inpainting model, we can accurately determine the placement of initial points, significantly elevating both the fidelity and efficiency of 3D Gaussian inpainting.  In particular, we first inpaint the depth in the reference view, then unproject the points into the 3D space to achieve optimal initialization. However, current depth inpainting methodologies~\cite{suvorov2022lama,ke2023repurposing,mirzaei2023spinnerf,ye2023gaussian} are often a limiting factor; commonly, they lack the generality required to accurately complete object depth, or they produce depth maps that misalign with the original, with errors amplified during unprojection. In this work, we harness the power of pre-trained latent diffusion models, training our depth inpainting model with diffusion-based priors to substantially enhance the quality of our inpainting results. The model exhibits a marked improvement in aligning with the unpainted regions and in reconstructing the depth of objects. This enhanced alignment capability ensures a more coherent extension of the existing geometry into the inpainted areas, leading to a seamless integration within the 3D scene. Furthermore, to address challenging scenarios involving large occlusions, we design \method with a progressive strategy that showcases its capability to resolve such complex cases.

%  What we achieve

Our extensive experiments on various datasets, which include both forward-facing and unbounded 360-degree scenes, demonstrate that our method outperforms the baseline approaches in terms of visual quality and inpainting speed, being 20 times faster. With the effective depth inpainting framework based on a pre-trained LDM,  we demonstrate that the integration of 3D Gaussians with depth inpainting offers an efficient and feasible approach to completing 3D scenes.  The strength of LDMs~\cite{rombach2022high,podell2023sdxl} is pivotal to our approach, allowing our model to inpaint not just the background but also to complete objects. Beyond the core functionality, our method facilitates additional applications, such as user-interactive texture inpainting, which enhances user engagement by allowing direct input into the inpainting process. We also demonstrate the adaptability of our method for downstream tasks, including scene manipulation and object insertion, revealing the broad potential of our approach in the context of editing and augmenting 3D spaces.
 
% \begin{itemize}
% \item Our proposed inpainting method significantly outperforms existing approaches on a variety of datasets, including both forward-facing and 360-degree scenes, excelling in terms of visual quality and editing speed.
% \item We have developed a straightforward yet effective depth inpainting framework based on a pre-trained LDM, which delivers compelling results on test sceneries. Furthermore, we demonstrate that the integration of 3D Gaussians with depth inpainting offers an efficient and feasible approach to completing 3D scenes.
% \item Our method enables further applications, such as user-interactive texture editing and object insertion, showcasing the versatility of the proposed inpainting technique.
% \end{itemize}

%% file: sections/2.related.tex
\section{Related Work}

\subsection{Image and Video Inpainting}
Image and video inpainting is an important editing task~\cite{xu2019deep,ouyang2023codef,zeng2020learning,newson2014video,sargsyan2023migan,ko2023continuously,chu2023rethinking,podell2023sdxl} that aims to restore the missing regions within an image or video by inferring visually consistent content. 
Traditional works for image inpainting~\cite{bertalmio2000image,ballester2001filling,tschumperle2005vector,efros1999texture,barnes2009patchmatch,darabi2012image} typically involve extracting low-level features to restore damaged areas.
Similarly, in video inpainting~\cite{wexler2007space,granados2012background,newson2014video,huang2016temporally,newson2013towards,shih2009exemplar,strobel2014flow}, the restoration process is often approached as an optimization task based on patch sampling.
However, these methods generally lack capacity when handling images with large missing regions or corrupted videos with complex motions.
Recently, deep learning has not only empowered inpainting models to overcome these challenges in restoration but has also expanded their capacity to generate new, semantically plausible content~\cite{quan2024deep}.
State-of-the-art image inpainting methods~\cite{lugmayr2022repaint,suvorov2022lama,li2022mat,dong2022incremental,yu2023inpaint,sargsyan2023migan,ko2023continuously,chu2023rethinking,podell2023sdxl} excel at effectively handling large mask inpainting tasks on high-resolution images;
cutting-edge techniques on video inpainting~\cite{zhou2023propainter,lee2023semantic,zheng2023ciri,xu2019deep,gao2020flow,zhang2022inertia,zhang2022flow,li2022towards,liu2021fuseformer,bertalmio2000image} commonly leverage flow-guided propagation and video Transformers to restore missing parts in videos with natural and spatiotemporally coherent content.

\subsection{3D Scene Inpainting}
With the increasing accessibility of 3D reconstruction models, there is a growing demand for 3D scene editing~\cite{cheng2023agap,zhuang2023dreameditor,haque2023instruct,pang2023locally,kobayashi2022decomposing,zhang2022arf,zhang2023refnpr}.
3D scene inpainting is one prominent application to fill in the missing parts within a 3D space, such as removing objects from the scene and generating plausible geometry and texture to complete the inpainted regions.
%
% The objective of 3D scene inpainting is to effectively fill in the missing parts within a 3D space, such as removing objects from the scene and generating plausible geometry and texture to complete the inpainted regions.
%
Early inpainting works mainly focuses on performing geometry completion~\cite{dai2020sgnn,dai2018scancomplete,kazhdan2006poisson,kazhdan2013screened,wu2015shapnets,dai2017shape,han2017high,wang2017shape,park2019deepsdf,song2017semantic}.
Recent advancements in 3D inpainting techniques have facilitated the simultaneous inpainting of both semantics and geometry by successfully handling the interplay between these two aspects~\cite{wei2023clutter}.
They can be broadly categorized into two groups based on the adopted 3D representation: NeRF~\cite{mildenhall2020nerf} and Gaussian Splatting (GS)~\cite{kerbl2023gaussian}.
Some NeRF-based methods~\cite{mirzaei2022laterf,kobayashi2022decomposing,kerr2023lerf,liu2023instance,siddiqui2023panoptic} leverage CLIP~\cite{radford2021clip} or DINO features~\cite{caron2021emerging} to learn 3D semantics for inpainting; 
others~\cite{liu2022nerfin,mirzaei2023spinnerf,weder2023removing,cheng2021rethinking,mirzaei2023reference,wang2023inpaintnerf360,weber2023nerfiller} typically rely on 2D image inpainting models with depth or segmentation priors to optimize NeRFs through neural fields rendering.
In contrast to inpainting on NeRF, several methods~\cite{ye2023gaussian,chen2023gaussianeditor,huang2023point,jiang2024vrgs} explore inpainting techniques on GS models, thanks to their notable advantages such as impressive rendering efficiency and high-quality reconstruction.
In our paper, we further improve the efficiency and the quality of 3D inpainting within GS settings.

\subsection{Diffusion Models for Monocular Depth}
The explicit nature of 3D Gaussians makes the accurate allocation of inpainted points within 3D scenes (\textit{e.g.}, object surfaces) highly beneficial for 3D scene inpainting via optimization.
A direct and effective solution is to utilize the 2D depth prior of reference views obtained through monocular depth estimation~\cite{eigen2015predicting,godard2017unsupervised,zhou2017unsupervised,li2018megadepth,ranftl2022towards,bhat2023zoedepth,yang2024depth} or completion~\cite{liu2017robust,zhang2018deep,lu2014depth,shih2020photography,besic2022dynamic,fujii2020rgbd,zhang2022indepth,makarov2021depth} models to initialize the inpainted 3D points.
Thanks to the superior performance of latent diffusion models (LDM)~\cite{dickstein2015deep,ho2020ddpm,song2021ddim,rombach2022high,podell2023sdxl,blattmann2023align,betker2023dalle3,saharia2022photorealistic}, it opens up the possibility of enhancing depth learning by leveraging or distilling the capabilities of these models.
Several methods have attempted to employ diffusion priors for estimating monocular depth~\cite{ke2023repurposing,ji2023ddp,duan2023diffusiondepth,saxena2023monocular,saxena2023the,zhao2023unleashing}.
However, learning from LDM for monocular depth completion (or inpainting) receives less attention.
While some methods~\cite{mirzaei2023spinnerf,weder2023removing} employ LaMa~\cite{suvorov2022lama} to inpaint depth in the Jet color space, the precision of the resulting inpainted depth map is compromised due to the lossy quantization process when converting metric depth to the Jet color space.
To the best of our knowledge, our work is the first resolve this problem by training an accurate depth completion model from diffusion prior~\cite{podell2023sdxl}.

% Apart from depth estimation, 

% from Diffusion Prior

% Several methods have tried to use DDPMs for metric depth
% estimation. The DDP approach [20] proposes an architecture to encode the image but decode a depth map and has
% obtained state-of-the-art results on the KITTI dataset. DiffusionDepth [10] performs diffusion in the latent space, conditioned on image features extracted with a SwinTransformer.
% DepthGen [42] extends a multi-task diffusion model to metric depth prediction, including handling noisy ground truth.
% Its successor DDVM [41] emphasizes pretraining on synthetic and real data for enhanced depth estimation. Finally,
% VPD [62] employs a pretrained Stable Diffusion model as
% an image feature extractor with additional text input.
% Our approach advances beyond these methods, which
% perform well but only in their specific training domains. We
% explore the potential of pretrained LDMs for single-image
% depth estimation across diverse, real-world settings.

%% file: sections/3.method.tex
\section{Method}

% \begin{figure*}[t]
%     \centering
%     \includegraphics[width=0.97\linewidth]{figures/progressive.pdf}
%     % \vspace{-9pt}
%     \caption{%
%         \textbf{Pipeline Placeholder}. 
%     }
%     \label{fig:prog}
%     % \vspace{-7pt}
% \end{figure*} 

\begin{figure*}[t]
    \centering
    \includegraphics[width=1\linewidth]{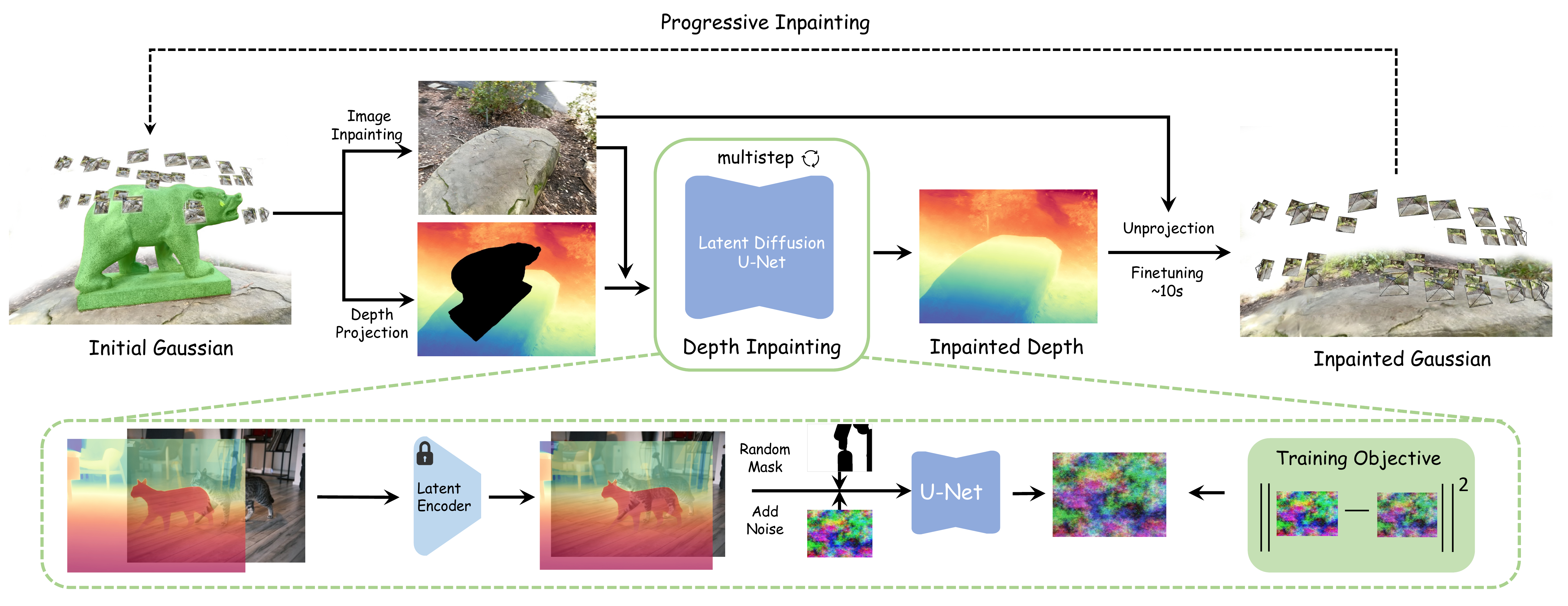}
    % \vspace{-9pt}
    \caption{%
        % \textbf{Pipeline}. 
    % {\bf An illustration of Infusion driven by Depth Inpainting}. Given a selected target to be removed on the top of the optimized 3D Gaussians, the proposed InFusion inpaints one-view RGB image and masked out the depth projection of 3D Gaussians to be null. Then, a depth inpainting model is leveraged to obtain an {\em inpainted depth} for the multi-view consistent 3D Gaussian inpainting. 
    % }
    % {\bf Illustration of Infusion driven by Depth Inpainting}. {\em Top}: Given a target to be removed from the optimized 3D Gaussians, our proposed InFusion method first inpaints an arbitrarily selected one-view RGB image and sets the depth projection of the targeted 3D Gaussians to be inpainted using the proposed diffusion model of depth inpainting. The view-dependent occlusion issue is addressed by the progressive scheme of depth inpainting, using the other unobstructed viewpoints. {\em Bottom}: A close-up illustration of the {\em training pipeline} of the depth inpainting U-Net, in which we leverage a mask-driven denoising diffusion training for the latent U-Net with the frozen latent tokenizer of both the RGB image and the input depth map.
    % }
    {\bf Illustration of Infusion driven by Depth Inpainting}.
    {\em Top}: To remove a target from the optimized 3D Gaussians, our \method  first inpaints a selected one-view RGB image and applies the proposed diffusion model for depth inpainting to the depth projection of the targeted 3D Gaussians. The progressive scheme addresses view-dependent occlusion issues by utilizing other unobstructed viewpoints.
    {\em Bottom}: A detailed view of the training pipeline for the depth inpainting U-Net is presented. We employ a mask-driven denoising diffusion for training of the U-Net, which utilizes a frozen latent tokenizer by taking the RGB image and depth map as inputs.
    }
    \label{fig:training_pipe}
    % \vspace{-7pt}
\end{figure*} 

\iffalse
\begin{figure*}[t]
    \centering
    \includegraphics[width=1\linewidth]{figures/method2.pdf}
    % \vspace{-9pt}
    \caption{%
        % \textbf{Pipeline}. 
        {\bf Training Pipeline of the Diffusion Model of Depth Inpainting}. By tokenizing the RGB image and the depth map using a frozen latent encoder, the proposed method first leverages the random masks in the concatenation while adding random noises. The noisy result is further concatenated with the reconstructed RGB image as the input of the U-Net for denoising diffusion. 
    }
    \label{fig:training_pipe}
    % \vspace{-7pt}
\end{figure*} 
\fi 
% \method utilizes two core modules for 3D Gaussian inpainting: the first module encompasses a diffusion model specifically designed for depth completion, and the second module employs a 3D Gaussian reconstruction procedure that leverages the depth predictions from the diffusion model to achieve seamless 3D Gaussian inpainting.

%
% We provide a concise overview of the problem formulation in \cref{problem_formulation}. Subsequently, we describe the details of the diffusion-based depth completion model in \cref{diff_depth} and how we use this model as a prior for 3D scene inpainting in \cref{gaussian_inpaint}.

\subsection{Overview}\label{problem_formulation}

% Given a collection of multi-view images $\mathcal{I}=\{I_i\}_{i=1}^{n}$, accompanied by respective camera poses $\Pi=\{\Pi_i\}_{i=1}^{n}$ and corresponding masks $\mathcal{M}=\{m_i\}_{i=1}^{n}$ which identifies the regions for inpainting, our objective is to construct a Gaussian Splatting (GS)~\cite{kerbl2023gaussian} model that synthesizes \textit{inpainted} images from arbitrary novel views.

% %%
% % However, 3D scene inpainting of Gaussian Splatting (GS)\tocite{} is a non-trivial task.
% %
% GS represents 3D scenes using point clouds where each Gaussian's ($\Theta$) property is characterized by a set of attributes: a centroid $x\in\mathbb{R}^3$, a color vector represented by spherical harmonics $c\in\mathbb{R}^k$, a scaling factor $s\in\mathbb{R}^3$, a rotation quaternion $q\in\mathbb{R}^4$, as well as an opacity scalar $\alpha$.
% %
% 3D Gaussian inpainting derives significant advantage from the precise allocation of inpainted points of 3D scenes(e.g., object surfaces).
% %
% Consequently, a direct and effective strategy involves executing depth completion from the selected reference views.
% %
% Depth information facilitates precise position for points, thereby enabling a well-informed initialization of the 3D Gaussians according to the completed depth.

% Building upon the aforementioned insights, we first devise a diffusion-based depth completion model conditioned on the color image. Leveraging this depth completion framework, we then proceed to execute 3D Gaussian inpainting.

% 

Formally, 3D scenes can be represented by 3D Gaussians $\Theta$ , given a collection of multi-view images $\mathcal{I}=\{I_i\}_{i=1}^{n}$, accompanied by respective camera poses $\Pi=\{\pi_i\}_{i=1}^{n}$~\cite{kerbl2023gaussian}. Our objective is to edit the scene $\Theta$ with a particular focus on inpainting, which aims to supplement an incomplete set of 3D Gaussians. The complexity of 3D Gaussian inpainting arises due to potential inconsistencies in the supervision provided by the 2D inpainted images from multiview. Nevertheless, three key observations inspire our solution design to address the challenges:

\begin{itemize}
\item The reconstruction quality of the optimized 3D Gaussians for novel view synthesis is highly sensitive to the initialization, especially when the view number is limited. Hence, we are motivated to carefully place the initial points within the inpainting regions for enhancing the inpainting quality.
% careful placement of the initial points within the inpainting regions is crucial for enhancing quality.

\item Contemporary research indicates that the initialization of 3D Gaussians with unprojected depth maps\cite{chung2023luciddreamer,ouyang2023text2immersion,charatan2023pixelsplat } yields promising results due to explicitness. This observation implies that using inpainted depth images for initializing the missing region could be advantageous.

\item Incorporating a diffusion prior~\cite{avrahami2022blended,blattmann2023align,podell2023sdxl,rombach2022high,ke2023repurposing} into depth estimation markedly improved accuracy especially for general objects. This finding indicates that a similar approach can be adopted to leverage diffusion priors for benefiting depth inpainting.

\end{itemize}

Leveraging the key observations discussed earlier, our pipeline is illustrated in Figure~\ref{fig:training_pipe}. Starting with the 3D Gaussians $\Theta$, we first segment out and discard unwanted Gaussians under the guidance of masks $\mathcal{M}=\{{m_i}\}_{i=1}^{n}$, which delineate the targeted regions for modification. As mentioned, depth inpainting can play a crucial role in determining the initial placement of Gaussians. To achieve this, we select a reference view and perform inpainting on both the image and its corresponding depth map to facilitate accurate unprojection. Existing depth inpainting models may not possess the versatility needed for precise depth completion or may produce depths that are inconsistent with the unpainted regions. Such misalignments lead to suboptimal inpainting outcomes. To address this, we develop a more generalized depth inpainting model that harnesses the strengths of natural diffusion processes.  In situations with substantial occlusion, relying on a single reference view may prove insufficient. To solve this, our approach incorporates multiple reference views through a progressive inpainting strategy.

The remainder of our methods are structured as the following. We describe the specifics of the diffusion-based depth completion model in \cref{diff_depth} and use this model to do 3D scene inpainting in \cref{gaussian_inpaint}. Finally, we provide the details of progressive inpainting in \cref{progressive}.

% The diffusion prior not only enhances the model's capabilities but also ensures more coherent and visually pleasing inpainting results.

%%
% However, 3D scene inpainting of Gaussian Splatting (GS)\tocite{} is a non-trivial task.
%
% GS represents 3D scenes using point clouds where each Gaussian's ($\Theta$) property is characterized by a set of attributes: a centroid $x\in\mathbb{R}^3$, a color vector represented by spherical harmonics $c\in\mathbb{R}^k$, a scaling factor $s\in\mathbb{R}^3$, a rotation quaternion $q\in\mathbb{R}^4$, as well as an opacity scalar $\alpha$.
% %
% 3D Gaussian inpainting derives significant advantage from the precise allocation of inpainted points of 3D scenes(e.g., object surfaces).
% %
% Consequently, a direct and effective strategy involves executing depth completion from the selected reference views.
% %
% Depth information facilitates precise position for points, thereby enabling a well-informed initialization of the 3D Gaussians according to the completed depth.

% Building upon the aforementioned insights, we first devise a diffusion-based depth completion model conditioned on the color image. Leveraging this depth completion framework, we then proceed to execute 3D Gaussian inpainting.

\subsection{Diffusion Models for Depth Completion}\label{diff_depth}
 A precise and reliable depth inpainting model is essential to obtain a well-founded set of initial points for inpainting Gaussians. We build our depth completion model on latent diffusion models (LDMs)~\cite{rombach2022high} for the strong priors due to their training on extensive, internet-scale collections of images. Given a set of color images and their corresponding depth, as well as various random masks, we seek to learn a model with the ability to inpaint the masked depth. The following three sections describe our diffusion-based depth completion model in details. 

\noindent \textbf{Diffusion Models} 
% We build our depth completion model on latent diffusion models (LDMs)~\tocite{} for their strong priors since they have been trained on internet-scale image collections specifically. LDMs perform diffusion steps in a low-dimensional latent space via a pretrained variational auto-encoder (VAE) $\mathcal{E}$. Diffusion steps are performed on these nosiy latents where a denosing U-Net $\epsilon_\theta$ iteratively removes noise to get clean latents. During inference, this U-Net is used to iteratively denoise pure Gaussian noise to a clean latent. To recover an image, the latents are passed through a VAE decoder $\mathcal{E}$.
% similar to Marigold~\cite{ke2023repurposing}
We formulate depth completion as a task of conditional denoising diffusion generation.  The LDMs operates by conducting diffusion processes within a lower-dimensional latent space, facilitated by a pre-trained Variational Auto-Encoder (VAE) $\mathcal{E}$. Diffusion steps are performed on these noisy latents where a denosing U-Net $\epsilon_\theta$ iteratively removes noise to get clean latents. During inference, the U-Net is applied to denoise pure Gaussian noise into a clean latent. The image recovery is then achieved by passing these refined latents through the VAE decoder $\mathcal{D}$. This ensures that the depth completion model benefits from the powerful generative capabilities inherent in LDMs while also maintaining efficiency by operating within a compressed latent space.

\noindent \textbf{Training}
We develop our model on top of a pre-trained text-to-image LDM (Stable Diffusion~\cite{rombach2022high}) to save computational resources and enhance training efficiency. Modifying the existing model architecture, we adapt it for image-conditioned depth completion tasks. An outline of the refined fine-tuning process is presented in \cref{fig:training_pipe}.

Our depth completion diffusion model accepts a trio of inputs: a depth map $d$, a corresponding color image $I$, and a mask $m$. Leveraging the frozen VAE, we encode both the color image and the depth map into a latent space, which serves as the foundation for training our conditional denoiser. To accommodate the VAE encoder's design for 3-channel (RGB) inputs when presented with a single-channel depth map, we duplicate the depth information across three channels to create an RGB-like representation.  We apply a linear normalization to ensure the depth values predominantly reside within the interval $[-1, 1]$ following Marigold~\cite{ke2023repurposing}, thereby conforming to the VAE's expected input range. This normalization is executed via an affine transformation delineated as follows:
\begin{equation}
    d'=\frac{d-d_2}{d_{98}-d_2}\times2 - 1,
\end{equation}
where $d_2$ and $d_{98}$ represent the $2^{nd}$ and $98^{th}$ percentiles of individual depth maps, respectively. Such normalization facilitates the model's concentration on affine-invariant depth completion, enhancing the robustness of the algorithm against scaling and translation.

The normalized depth $d'$ and the color image are first encoded into the latent space with the encoder of the VAE:
\begin{equation}
    z^{(d')}=\mathcal{E}(d'), \\
    z^{(I)}=\mathcal{E}(I),
\end{equation}
The encoder produces a 4-channel feature map that has a lower resolution than the original input. To construct the image-conditioned depth completion model, we initially resize the mask $m$ to align with the dimensions of $z^{(d')}$, yielding $m' = \text{downsample}(m)$. We then create a composite feature map by concatenating the noisy latent depth code $z^{(d')}_t$, the element-wise product of the clean latent depth code and the downscaled mask $z^{(d')}_m = z^{(d')} \odot m'$, and the latent image code $z^{(I)}$, along with $m'$, as follows:
\begin{equation}
z_t = \text{cat}(z^{(d')}_t, z^{(d')}_m, z^{(I)}, m'), \label{equ:cat}
\end{equation}
along the channel dimension, where $z^{(d')}_t=\alpha_t z^{(d')}+\sigma_t\epsilon$. The concatenated feature map $z_t$, comprising $4+4+4+1=13$ channels, is subsequently fed into the U-Net-based denoiser $\epsilon_{\theta}$.

At training time, U-Net parameters $\theta$ are updated by taking a data pair $(I, d, m)$ from the training set, noising $d$ with sampled noise $\epsilon$ at a random timestep $t$, computing the noise estimate $\hat{\epsilon}=\epsilon_{\theta}(z_t)$ and minimizing the denoising diffusion objective function following DDPM~\cite{ho2020ddpm}:
\begin{equation}
    \mathcal{L}=\mathbb{E}_{d, \epsilon, t}\lVert \epsilon-\epsilon_{\theta}(z_t) \rVert_2^2,
\end{equation}
where $t\in\{1,2,...,T\}$ indexes the diffusion timesteps, $\epsilon\in\mathcal{N}(0, I)$, and $z_t$, the noisy latent at timestep $t$, is calculated as \cref{equ:cat}.

\noindent \textbf{Inference}
The inference of our depth completion model commences with an input comprising a depth map $d$, its corresponding color image $I$, and a mask $m$ that delineates the target completion region. The color image $I$ undergoes SDXL-based~\cite{podell2023sdxl} image inpainting, resulting in $\tilde{I} = \mathcal{F}_I(I, m)$, where $\mathcal{F}_I$ represents the image inpainting model. Subsequently, we generate the concatenated feature map as defined in \cref{equ:cat}, which is then progressively refined according to the fine-tuning scheme. Leveraging the non-Markovian sampling strategy from DDIM~\cite{song2021ddim} with re-spaced steps facilitates an accelerated inference. The final depth map is then derived from the latent representation decoded by the VAE decoder $\mathcal{D}$, followed by channel-wise averaging for post-processing.

% Different from Marigold, our method eschews the use of ensembling during inference. Our objective is to complete the masked region of a given depth map based on the corresponding color image, rather than estimating the full depth map from the color image alone. This approach results in lower uncertainty, yielding a more stable inference process and obviating the need for ensembling to aggregate multiple inference outputs.
% We begin by fitting an impaired 3D Gaussians $\Theta$ excluding the mask region with the multi-view images set $\mathcal{I}=\{I_i\}_{i=1}^{n}$, camera poses $\Pi=\{\Pi_i\}_{i=1}^{n}$, and their corresponding masks $\mathcal{M}=\{m_i\}_{i=1}^{n}$ using the masked version of the $\mathcal{L}_1$ and D-SSIM loss:
% \begin{equation}
%     \mathcal{L}=(1-\lambda)\frac{1}{n}\sum_{i=1}^{n}\lVert (I'_i - I_i)\odot m_i \rVert_1 + \lambda\frac{1}{n}\sum_{i=1}^{n}\text{SSIM}(I'_i\odot m_i, I_i\odot m_i),
% \end{equation}
% where $I'_i$ denotes the GS-rendered image for the $i^{th}$ view.

%

\subsection{Inpainting 3D Gaussians with Diffusion Priors}\label{gaussian_inpaint}
The trained diffusion model generates plausible depth completions, thereby serving as an effective initialization for the 3D Gaussians. Upon removing undesired points from 3D Gaussians, a set of reference views $\{I_{s(i_j)}\}_{j=1}^r$ is selected, where $s(i_j) \in \{1,2,...,n\}$ and $r$ denotes the total number of chosen views. For forward-facing and certain 360-degree inward-facing datasets, a single reference view ($r=1$) is usually sufficient, whereas for more complex 360-degree scenes with occlusions, multiple reference views ($r>1$) are required. In instances with $r>1$, a progressive inpainting strategy is employed, detailed further in \cref{progressive}. The current discussion is focused on the $r=1$ scenario.

Assuming without loss of generality, for $r=1$, we designate the $s(i_1)^{th}$ view as the single reference. Initially, the color image $I_{s(i_1)}\odot m_{s(i_1)}$ is inpainted using an SDXL-based inpainting model to yield the restored image $\tilde{I}_{s(i_1)}$. The depth for the $s(i_1)^{th}$ view is then determined analogous to color rendering in GS:
\begin{equation}
d_{s(i_1)} = \sum_{i\in N_{s(i_1)}}z_i\alpha_i\prod_{j=1}^{i-1}(1-\alpha_j),
\end{equation}
where $z_i$ denotes the z-coordinate in the world coordinate system, and $\alpha_i$ represents the density of the corresponding point. It is important to note that the resulting depth $d$ is incomplete, as it derives from $\Theta$. To address this, we apply our diffusion-based depth completion model $\mathcal{F}_d$, which produces the refined depth map:
\begin{equation}
\tilde{d}_{s(i_1)} = \mathcal{F}_d(d_{s(i_1)}, \tilde{I}_{s(i_1)}, m_{s(i_1)}).
\end{equation}
With the completed depth map $\tilde{d}_{s(i_1)}$, the inpainted image $\tilde{I}_{s(i_1)}$, and the corresponding camera pose $\Pi_{s(i_1)}$, we unproject $\tilde{d}_{s(i_1)}$ and $\tilde{I}_{s(i_1)}$ from image space to 3D coordinates to form a colored point cloud $\mathcal{P}_{s(i_1)}$. This point cloud is then merged with the original 3D Gaussian point cloud to achieve a robust initialization $\Theta'$ for subsequent GS fine-tuning.

Ultimately, the preliminary Gaussian model $\Theta'$ are fine-tuned merely 50$\sim$150 iterations to yield the final Gaussian model $\tilde{\Theta}$, using solely the selected view image $I_{s(i_1)}$. The optimization is also guided by $\mathcal{L}_1$ combined with D-SSIM at the $s(i_1)^{th}$ view:
\begin{equation}
\mathcal{L}_{s(i_1)} = (1-\lambda)\lVert I'_{s(i_1)} - \tilde{I}_{s(i_1)} \rVert_1 + \lambda\cdot\text{D-SSIM}(I'_{s(i_1)}, \tilde{I}_{s(i_1)}),
\end{equation}
where $I'_{s(i_1)}$ denotes the image rendered from the $s(i_1)^{th}$ viewpoint. We set $\lambda=0.2$ across all experiments and provide comprehensive details of the learning schedule and additional experimental settings in~\cref{exp_set}.

\subsection{Progressive Inpainting}\label{progressive} 
For occlusion-rich, complex scenes, multiple reference views ($r>1$) are imperative. To solve these challenges, we implement a progressive inpainting approach. Commencing with the initial reference view $s(i_1)$ from the selected views $\mathcal{S}=\{s(i_1), s(i_2), ..., s(i_r)\}$, we apply Gaussian inpainting as delineated in \cref{gaussian_inpaint}. Subsequent to this, we render the color image, depth map, and associated mask from the next reference view $s(i_2)$. This process is iterated, employing Gaussian inpainting for each successive reference view until the view $s(i_r)$ is addressed. This progressive technique effectively accommodates the complexities, especially for occlusions.

%% file: sections/4.exp.tex
\section{Experiments}

% Summarize as follows:

\subsection{Experiments Setup}\label{exp_set}

\begin{figure*}[t]
    \centering
    \includegraphics[width=0.97\linewidth]{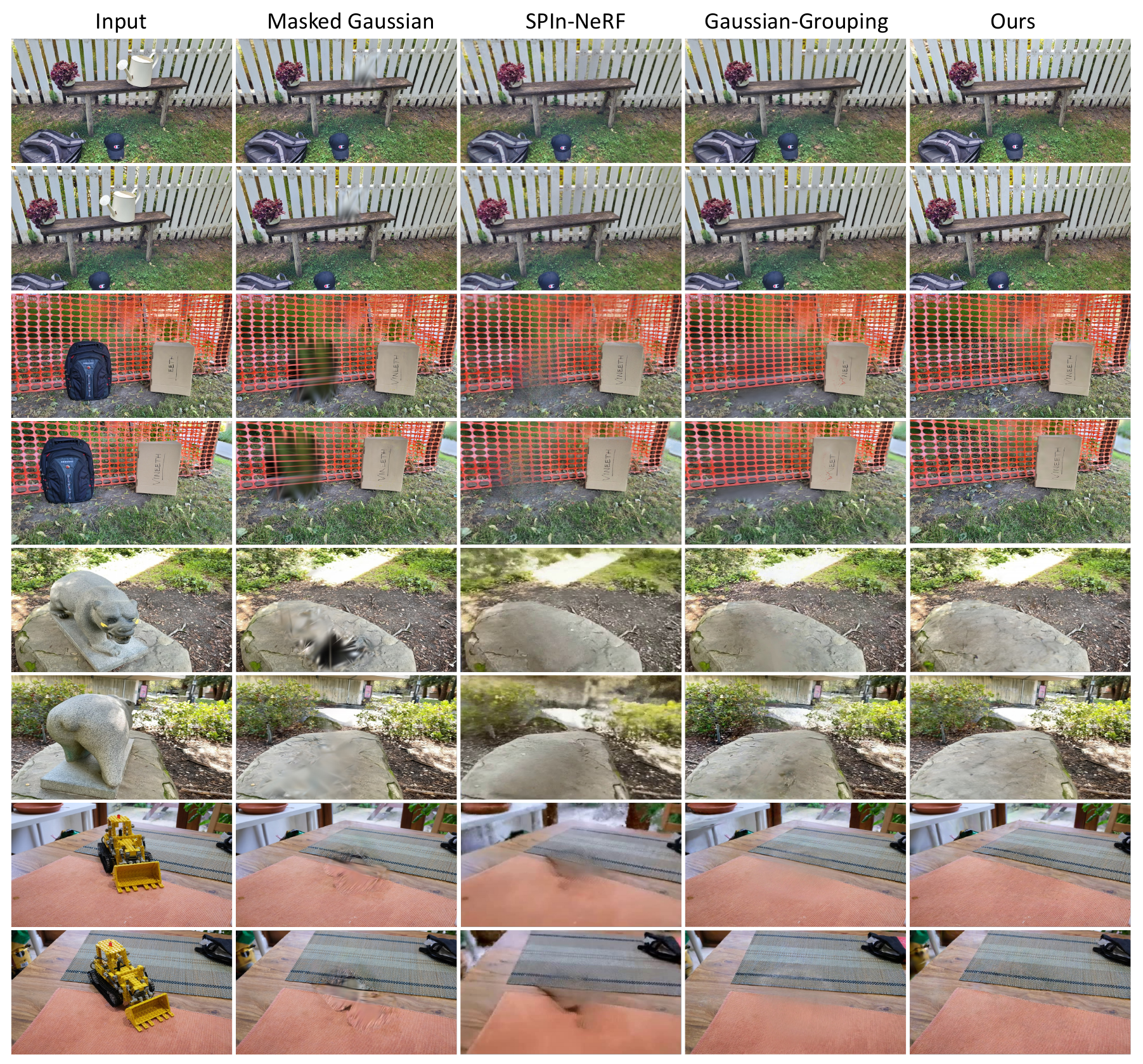}
    % \vspace{-9pt}
    \caption{%
        \textbf{Qualitative Comparison with Baselines}. Zoom in for details.  Our method exhibits sharp textures that maintain 3D coherence, whereas baseline approaches often yield details that appear blurred.
    }
    \label{fig:qualitative}
    % \vspace{-7pt}
\end{figure*} 

\begin{figure*}[t]
    \centering
    \includegraphics[width=0.97\linewidth]{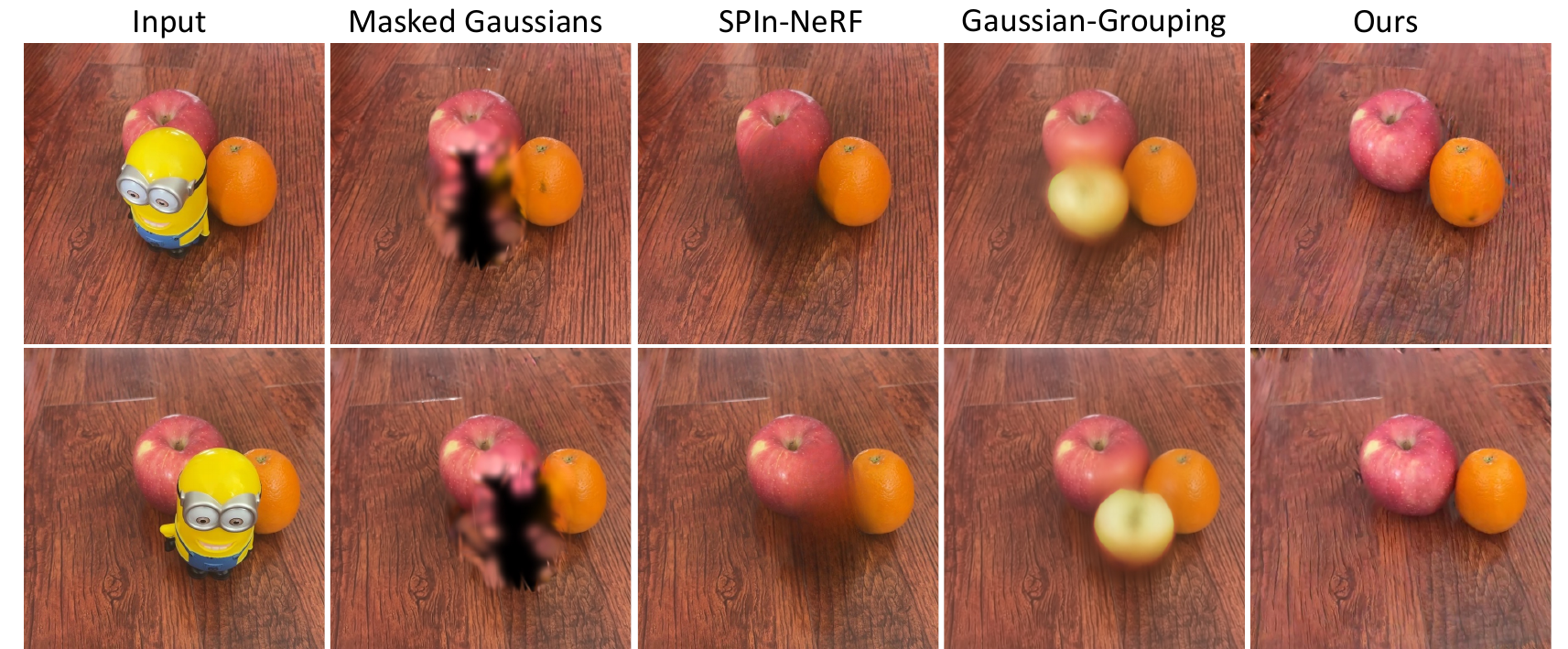}
    % \vspace{-9pt}
    \caption{%
        \textbf{Qualitative Comparison with Baselines}. We delve into more challenging scenarios, including those with multi-object occlusion, where our method uniquely stands out by accurately inpainting the obscured missing segments.
    }
    \label{fig:qualitative_difficult}
    % \vspace{-7pt}
\end{figure*} 

\begin{figure*}[t]
    \centering
    \includegraphics[width=0.97\linewidth]{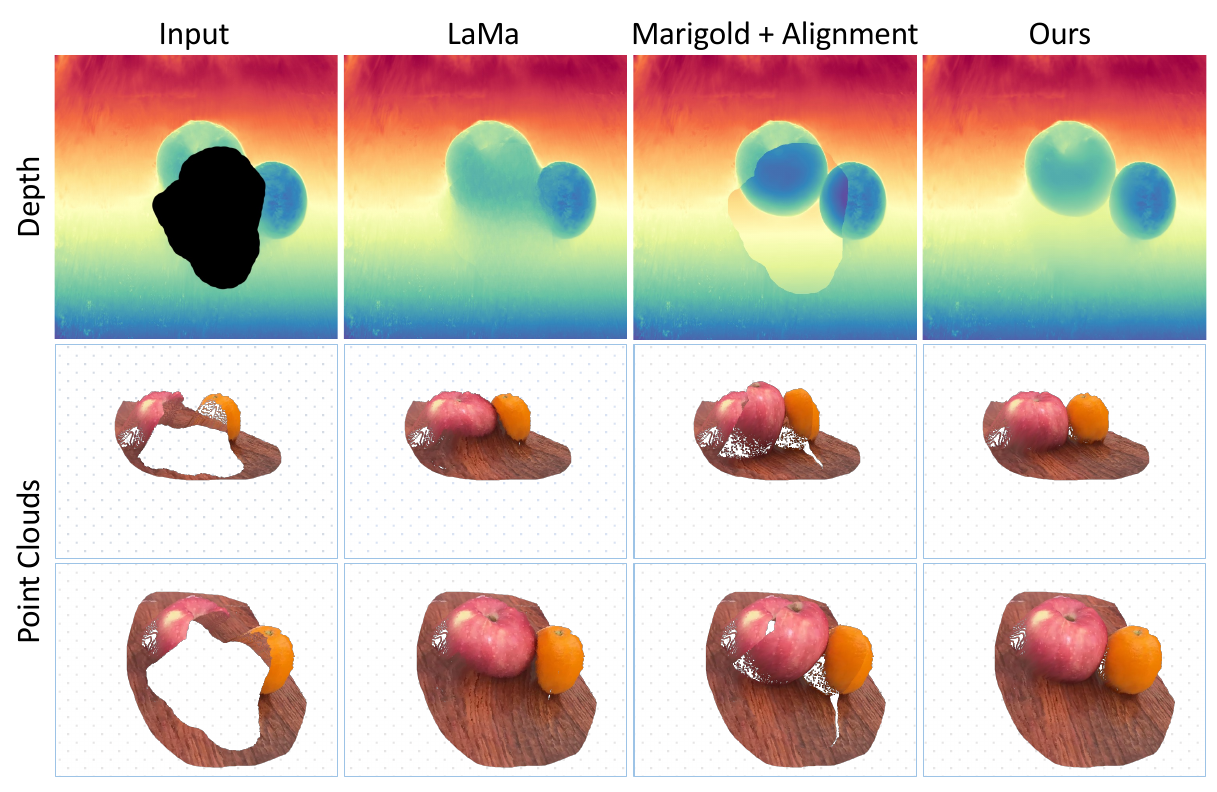}
    % \vspace{-9pt}
    \caption{%
        \textbf{Ablation study} on depth inpainting, we present comparative results against widely-used other baselines, along with the corresponding point cloud visualizations. The comparisons distinctly reveal that our approach successfully inpaints shapes that are correctly aligned with the existing geometry.
    }
    \label{fig:depth_ablation}
    % \vspace{-7pt}
\end{figure*}

\begin{figure*}[t]
    \centering
    \includegraphics[width=0.97\linewidth]{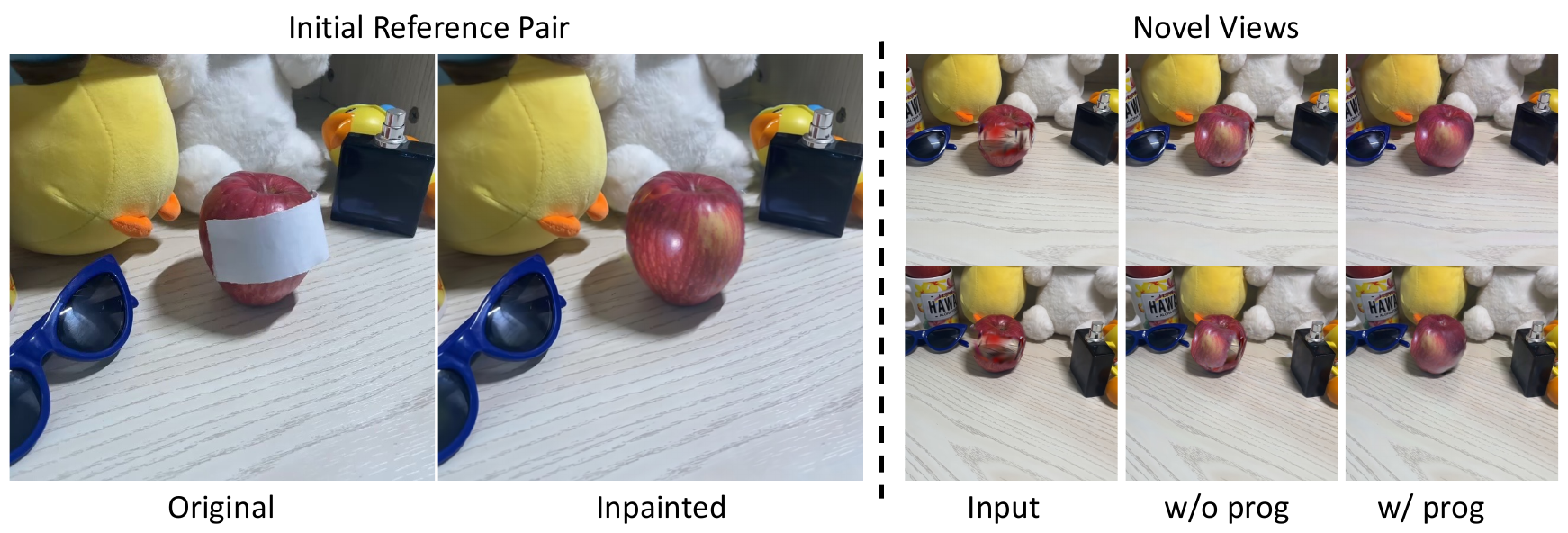}
    % \vspace{-9pt}
    \caption{%
        \textbf{Ablation study} on  progressive inpainting. \method can adeptly handle inpainting tasks for views that substantially deviate from the initial reference frames.
    }
    \label{fig:prog}
    % \vspace{-7pt}
\end{figure*}

\begin{figure*}[t]
    \centering
    \includegraphics[width=0.97\linewidth]{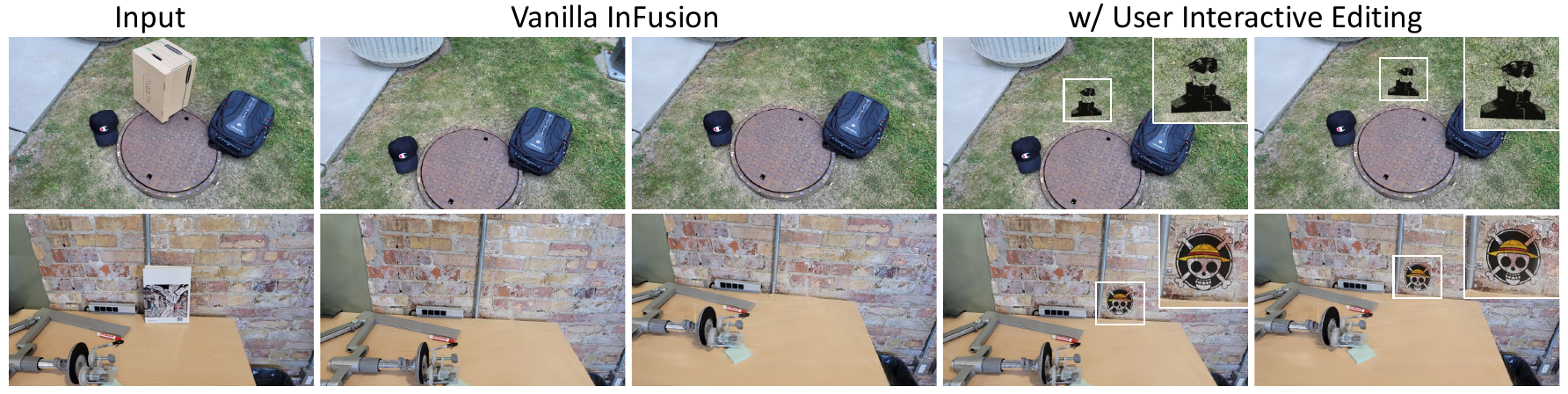}
    % \vspace{-9pt}
    \caption{%
        \textbf{User-interactive Texture Inpainting}.  \method allows users to modify the appearance and texture of targeted areas with ease.}
    \label{fig:app_0}
    % \vspace{-7pt}
\end{figure*}

\begin{figure*}[t]
    \centering
    \includegraphics[width=0.97\linewidth]{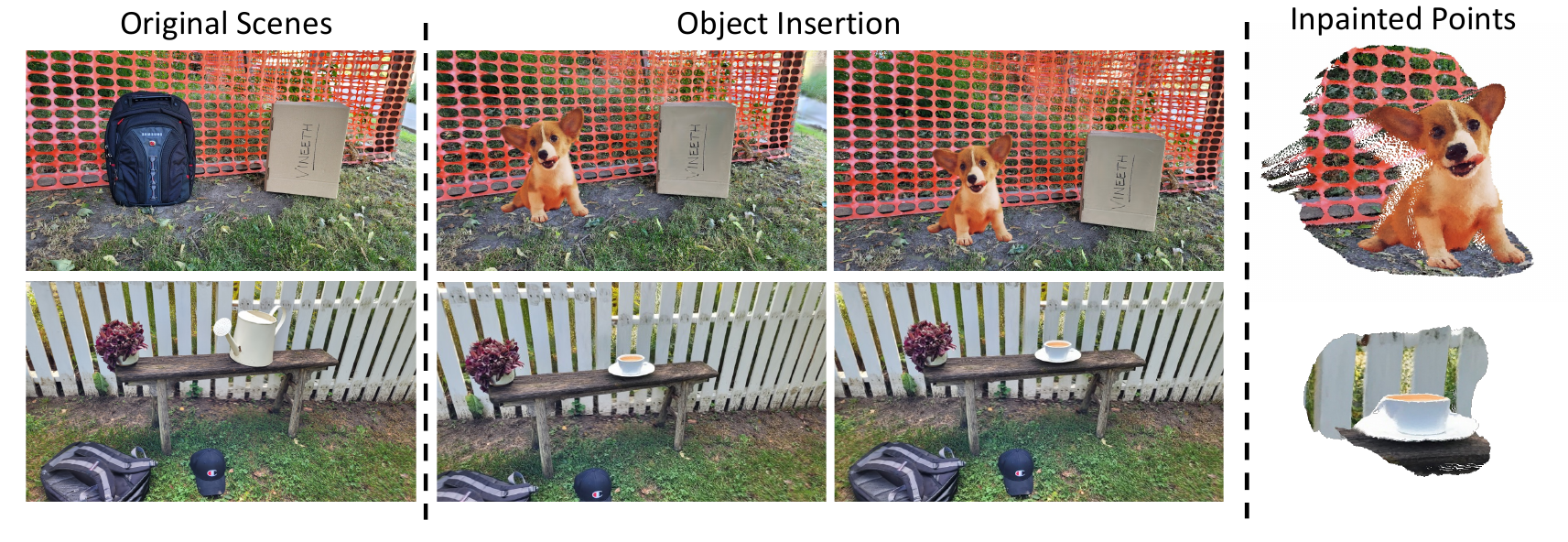}
    % \vspace{-9pt}
    \caption{%
        \textbf{Object Insertion.} Through editing a single image, users are able to project objects into a real three-dimensional scene. This process seamlessly integrates virtual objects into the physical environment, offering an intuitive tool for scene customization.
    }
    \label{fig:app_1}
    % \vspace{-7pt}
\end{figure*}

\begin{figure*}[t]
    \centering
    \includegraphics[width=0.97\linewidth]{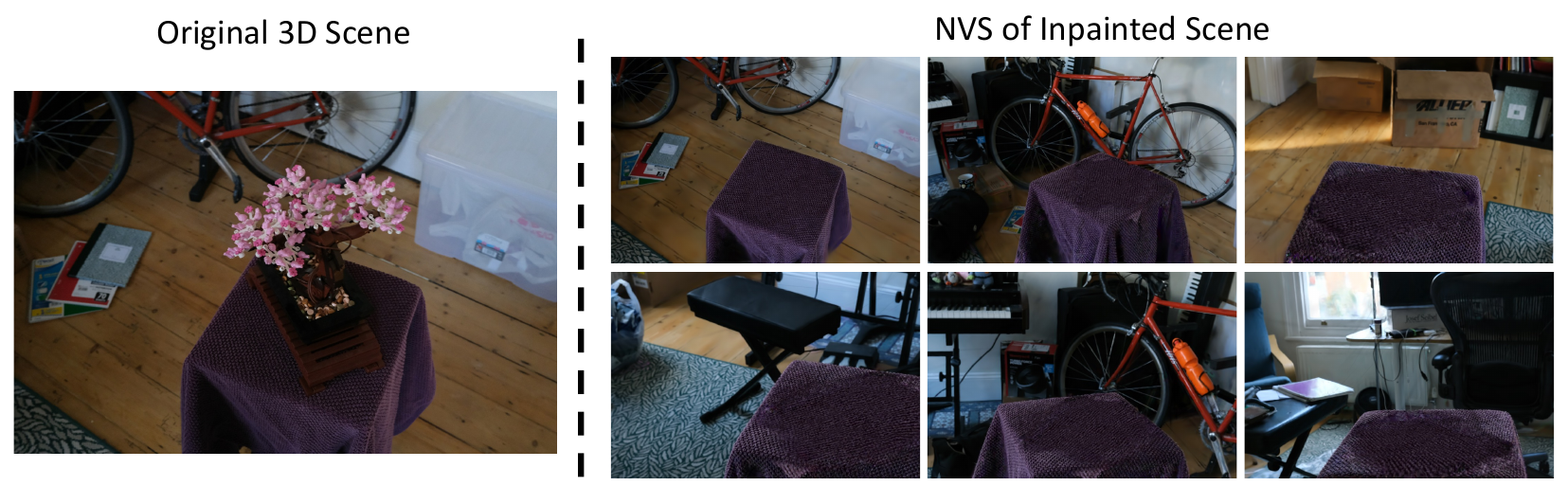}
    % \vspace{-9pt}
    \caption{%
        \textbf{Limitations}. As the lighting of the surrounding region increasingly differs from the reference, the inpainted area becomes less harmonious with these views. \method struggles to adapt inpainted regions to variations in lighting conditions.
    }
    \label{fig:limitation}
    % \vspace{-7pt}
\end{figure*}

\noindent \textbf{Training settings} To train a diffusion model with broad generalizability for depth inpainting and to facilitate generalized Gaussian inpainting, we train our LDM models using the SceneFlow dataset~\cite{mayer2016large}, which comprises FlyingThings and Driving scenes. This dataset offers an extensive collection of over 100,000 frames, each accompanied by ground truth depth, and rendered from a variety of synthetic sequences. During training, masks are randomly generated for each iteration using either a square, random strokes, or a combination of both techniques. We initialize the LDM with pre-trained depth prediction weights sourced from the Marigold~\cite{ke2023repurposing}. We also tested other pre-trained weight which is presented in the supplementary. The training process spans 200 epochs, with an initial learning rate of 1e-3, which is scheduled to decay after every 50 epochs. Utilizing eight A100 GPUs, the training process is completed within one day.

\noindent \textbf{Evaluation settings} We evaluate our method across a variety of datasets, which include forward-facing scenes and the more complex unbounded 360-degree scenes. For the forward-facing datasets, we adhere to the rigorous evaluation settings established by SPIn-NeRF~\cite{mirzaei2023spinnerf}. To further demonstrate the text-guided 3D inpainting capabilities of our method, we also introduce our own captured sequences including large occlusion between objects. The challenging unbounded 3D scenes are taken from the Mip-NeRF~\cite{barron2021mip}, featuring large central objects within realistic backgrounds, and the 3DGS, which includes a variety of intricate objects from free-moving camera angles. These datasets are particularly challenging for 3D inpainting. We emphasize that our LDM depth inpainting methods were not trained on any of these datasets. For scene masking, we used masks from SAM-Track~\cite{cheng2023segment}, dilating them by 9 pixels to reliably remove any undesired parts.

\noindent \textbf{Baseline selection} Our method has been benchmarked against three distinct baselines, each representing a different approach to inpainting. Spin-NeRF~\cite{mirzaei2023spinnerf} stands out as one of the best NeRF-based methods for 3D inpainting, offering 3D-aware results. For techniques leveraging 3D Gaussians, Gaussian-grouping~\cite{ye2023gaussian} is the state-of-the-art, building on the InstructNeRF2NeRF~\cite{haque2023instruct} framework to incorporate a pre-trained diffusion model for the inpainting task.
%Additionally, we evaluate against Propainter, a cutting-edge 2D video inpainting method. 
Lastly, we include a critical baseline where we forgo the depth diffusion inpainting process and instead directly optimize 3D scenes using the inpainted reference image with the aid of Stable Diffusion XL~\cite{podell2023sdxl}.

\subsection{Results Comparison} 
We conduct comparison with baseline methods in several aspects. For our quantitative analysis, we evaluate based on two key metrics: \textbf{ Image quality} and \textbf{ Speed}. Image quality is evaluated following the previous works; we report the LPIPS and FID scores for inpainted scenes following the settings in SPInNerf~\cite{mirzaei2023spinnerf}. As detailed in \cref{tab:Quantitative}, our method outperforms the baseline techniques in both metrics, achieving the best scores. In terms of speed, our method demonstrates a significant advantage. Thanks to the precision of our initial inpainted points and the efficiency of our fine-tuning process—which necessitates only a minimal number of iterations (around 100)—our approach is considerably faster than baseline methods. 

\cref{fig:qualitative} illustrates a side-by-side comparison of the inpainted results and corresponding novel views generated by our method against those from baseline methods. While baselines are capable of reconstructing the broad outlines of missing regions, they often yield textures that lack sharpness. Our approach, on the other hand, consistently produces fine-detailed textures across all views.  Moreover, as shown in~\cref{fig:qualitative_difficult}, our method can handle more difficult cases which include multi-object occlusion.

% Given the limitations of static images in conveying 3D consistency, we strongly encourage readers to view the supplementary videos for a more dynamic and thorough comparison.

\begin{table}[!h]
    \vspace{-10pt}
    \centering
    % \scriptsize
    \SetTblrInner{rowsep=1.2pt}      % Row space.
    \SetTblrInner{colsep=4.0pt}      % Col space.
    \begin{tblr}{
        cells={halign=c,valign=m},   % Text alignment for all cells.
        column{1}={halign=l},        % Text alignment for the first column.
        hline{1,2,7}={1-6}{},        % Horizontal lines.
        hline{1,5}={1.0pt},          % Horizontal line width.
        vline{2}={1-6}{},          % Vertical lines.
        row{6}={bg=lightgray!50},
    }
    \                   &  Masked Gaussians & SPIn-NeRF     &  Gaussians Grouping  &  Ours   \\
     LPIPS $\downarrow$   & 0.594   & 0.465      & 0.454  & \textbf{0.421}         \\
     FID  $\downarrow$   & 278.32    &  156.64       & 123.48  & \textbf{92.62}         \\
     Time $\downarrow$  & 20min    & 5h & 20min  & \textbf{40s}  \\
    \end{tblr}
    \vspace{10pt}
    \caption{\textbf{Quantitative evaluation.} We conducted a quantitative evaluation of 3D inpainting techniques on the inpainted areas of held-out views from the SPIn-NeRF dataset.
    Our method achieves optimal results in perceptual metric (LPIPS) and feature-based statistical distance (FID). Additionally, our method significantly reduces the optimization time compared to previous methods.
    }
    \label{tab:Quantitative}
    % \raggedright\scriptsize{~$\boldsymbol{-}$~~: [14] not inherently supports local stylization but possible (depends on prompts)}\\
    % \raggedright\scriptsize{~$\dagger$~~: All methods have on par pre-training and rendering time.}
    % \vspace{-14pt}
\end{table}
% The improvement is not limited to textures alone; depth visualization reveals that shapes inpainted by our method exhibit a smoothness superior to that of the baselines. 

Furthermore, we extend our comparative analysis to include additional capabilities of our method relative to the baselines. Owing to our method's utilization of the reference image, it inherently supports interactive 3D texture inpainting—an operation the baseline methods cannot accommodate. Additionally, our LDM-based approach facilitates object completion in forward-facing scenes. These advanced functionalities are exemplified in the application section of this paper.

\subsection{Ablation Study}
In order to validate the components within our pipeline, we conducted a series of ablation studies on the key design elements.

\noindent \textbf{Depth Inpainting}: A common approach prior to our work involved using image-based inpainting methods, such as LaMa or SDXL Inpainting, for depth inpainting~\cite{mirzaei2023spinnerf, weder2023removing}. However, because of the domain gap and the model capability, the inpainted results are less accurate as in \cref{fig:depth_ablation} and \cref{fig:teaser}.  Another stream involves using monocular depth estimation followed by depth alignment~\cite{fang2023gaussianeditor}. As depicted in \cref{fig:depth_ablation}, this method often results in depth discontinuities within the inpainted regions, leading to misalignment with the scene's original depth. While depth alignment techniques can mitigate this error, significant discrepancies persist. 

% Consequently, the absence of our trained depth inpainting model leads to suboptimal novel view synthesis due to inaccurately placed initial points.

% \noindent \textbf{LDM Training Design}: The significance of initializing our LDM with a pre-trained model is highlighted through an ablation study where the model is randomly initialized. This study demonstrates a noticeable decline in performance, particularly with novel objects, underscoring the benefits derived from a pre-trained model's exposure to diverse natural image datasets for depth inpainting tasks.

\noindent \textbf{Progressive InFusion}: Progressive design has a direct impact on performance for difficult cases. As evidenced in our results, augmenting the number of views enhances the handling of occlusions (\cref{fig:prog}). Nonetheless, this boost in performance comes at the expense of increased inference time. In simpler scenes, where the task is to remove the outermost object, utilizing a single reference view suffices.

\subsection{Applications} \label{application}
To showcase the practical utility of our proposed method, we present two key downstream applications:

\noindent \textbf{Interactive Texture Editing}
Our framework facilitates user-interactive texture editing within inpainted regions by allowing modifications to the reference image. As illustrated in \cref{fig:app_0}, users can seamlessly integrate custom text into 3D scenes, enhancing the personalization of the 3D environment.

\noindent \textbf{Object Insertion} Leveraging our diffusion-based depth inpainting approach, we enable effortless object insertion within frontal-face scenes, as depicted in \cref{fig:app_1}. This capability extends to the insertion of user-selected objects into the inpainted 3D scenes, offering a versatile tool for scene customization. 

\subsection{Limitation}
While the proposed method achieves impressive results in 3D inpainting, it encounters two main limitations: first, in scenarios with significant lighting changes across various angles, the inpainted sections can struggle to integrate flawlessly with adjacent areas, as highlighted in \cref{fig:limitation}; second, the method falls short in text-guided inpainting of highly complex objects within 360-degree scenes, limited by the current consistency of inpainting models.

%% file: sections/5.conclusion.tex
\section{Conclusion}
In conclusion, our proposed methodology, \method, effectively delivers high-quality and efficient inpainting for 3D scenes using Gaussian models. Our evaluations, both quantitative and qualitative, attest to its performance and ease of use. Moreover, we demonstrate that incorporating diffusion priors significantly enhances our depth inpainting model. We are confident that this improved depth inpainting model holds promise for a variety of 3D applications, particularly in the realm of novel view synthesis. However, our method currently has limitations in handling variations in lighting and reconstructing highly complex structured objects. Despite these challenges, our approach forges a connection between LDM and 3D scene editing. This synergy harbors significant potential for future advancements and optimizations.

%% file: sections/6.ref.tex
\bibliographystyle{splncs04}
\bibliography{ref.bib}

%% file: supp_1.tex
% \documentclass[runningheads]{llncs}

% \usepackage[review]{eccv}
% % \usepackage{eccv}  % For camera-ready version
% % \usepackage[mobile]{eccv}

% \usepackage{eccvabbrv}
% \usepackage{graphicx}
% \usepackage[accsupp]{axessibility}  % Improves PDF readability for those with disabilities.
% \usepackage{capt-of,xcolor,xspace,enumitem}
% \usepackage{amsmath,amssymb,amsbsy,amsfonts,dsfont,pifont,bm,bbm,mathrsfs,mathtools,nicefrac}
% \usepackage{algorithm,algpseudocode,listings}
% \usepackage{booktabs,multirow,adjustbox,diagbox,threeparttable,tabularray}

% \usepackage[pagebackref,breaklinks,colorlinks,citecolor=eccvblue]{hyperref}
% % \usepackage{hyperref}  % For camera-ready version

% \usepackage{orcidlink}
% \usepackage{wrapfig}
% \usepackage[capitalize]{cleveref}  % Should be loaded after 'hyperref', and works perfectly with 'subfigure'.
\crefname{section}{Sec.}{Secs.}
\Crefname{section}{Section}{Sections}
\crefname{table}{Tab.}{Tabs.}
\Crefname{table}{Table}{Tables}
\crefname{figure}{Fig.}{Figs.}
\Crefname{figure}{Figure}{Figures}
\crefname{equation}{Eq.}{Eqs.}
\Crefname{equation}{Equation}{Equations}
\hyphenpenalty=1200

% \newcommand{\todo}[1]{{\color{red} [TODO: #1]}}
% \newcommand{\tocheck}[1]{{\color{red} [CHECK: #1]}}
% \newcommand{\tofinish}[1]{{\color{red} [FINISH: #1]}}
% \newcommand{\tomodify}[1]{{\color{red} [MODIFY: #1]}}
% \newcommand{\tocite}[1]{{\color{red} [TO CITE]}}
% \newcommand{\method}{{\texttt{InFusion}}\xspace}
% \newcommand{\supp}{\textit{Supplementary Material}\xspace}
% \newcommand{\yj}[1]{{\color{cyan} [Yujun: #1]}}

% \begin{document}
\newpage
% \title{Appendix}
% \titlerunning{Abbreviated paper title}

% \author{
%     First Author\inst{1}\orcidlink{0000-1111-2222-3333} \and
%     Second Author\inst{2,3}\orcidlink{1111-2222-3333-4444} \and
%     Third Author\inst{3}\orcidlink{2222--3333-4444-5555}
% }
% \authorrunning{F.~Author et al.}

% \institute{
%     Princeton University, Princeton NJ 08544, USA \and
%     Springer Heidelberg, Tiergartenstr.~17, 69121 Heidelberg, Germany
%     \email{lncs@springer.com}\\
%     \url{http://www.springer.com/gp/computer-science/lncs} \and
%     ABC Institute, Rupert-Karls-University Heidelberg, Heidelberg, Germany\\
%     \email{\{abc,lncs\}@uni-heidelberg.de}
% }

% \maketitle
\renewcommand\twocolumn[1][]{#1}
% \maketitle
\setcounter{section}{0}
\setcounter{figure}{0}

The appendix are structured as follows: We begin with a detailed description of our implementation, including the specifics of our training configurations and the outlier removal process for the point cloud. Subsequently, we undertake an in-depth examination of various issues discussed within the paper. To conclude, we provide a broader set of results, encompassing a wider array of scenes and viewpoints.

\section{Implementation Details} 
\subsection{Training details}\label{training_details} 
The Depth inpainting model is initialized with the Marigold~\cite{ke2023repurposing} weights. The architecture of the neural network is consistent with that of Stable Diffusion v1.5~\cite{rombach2022high}, with the exception of the first convolutional layer. Moreover, during both training and inference phases, the input to the text encoder is persistently an empty string. The UNet has 9 additional input channels (4 for the encoded masked-depth, 4 for the guided encoded image and 1 for the mask itself) whose weights were zero-initialized. During training, we generate synthetic masks and, in 30\% mask everything. In the context of data processing, we maintain the original aspect ratio of the images during both the training and inference stages, resizing them to a maximum resolution of 768 pixels on the longest side.
% \vspace{-50pt}
\begin{figure*}[h]
    \centering
    \includegraphics[width=0.9\linewidth]{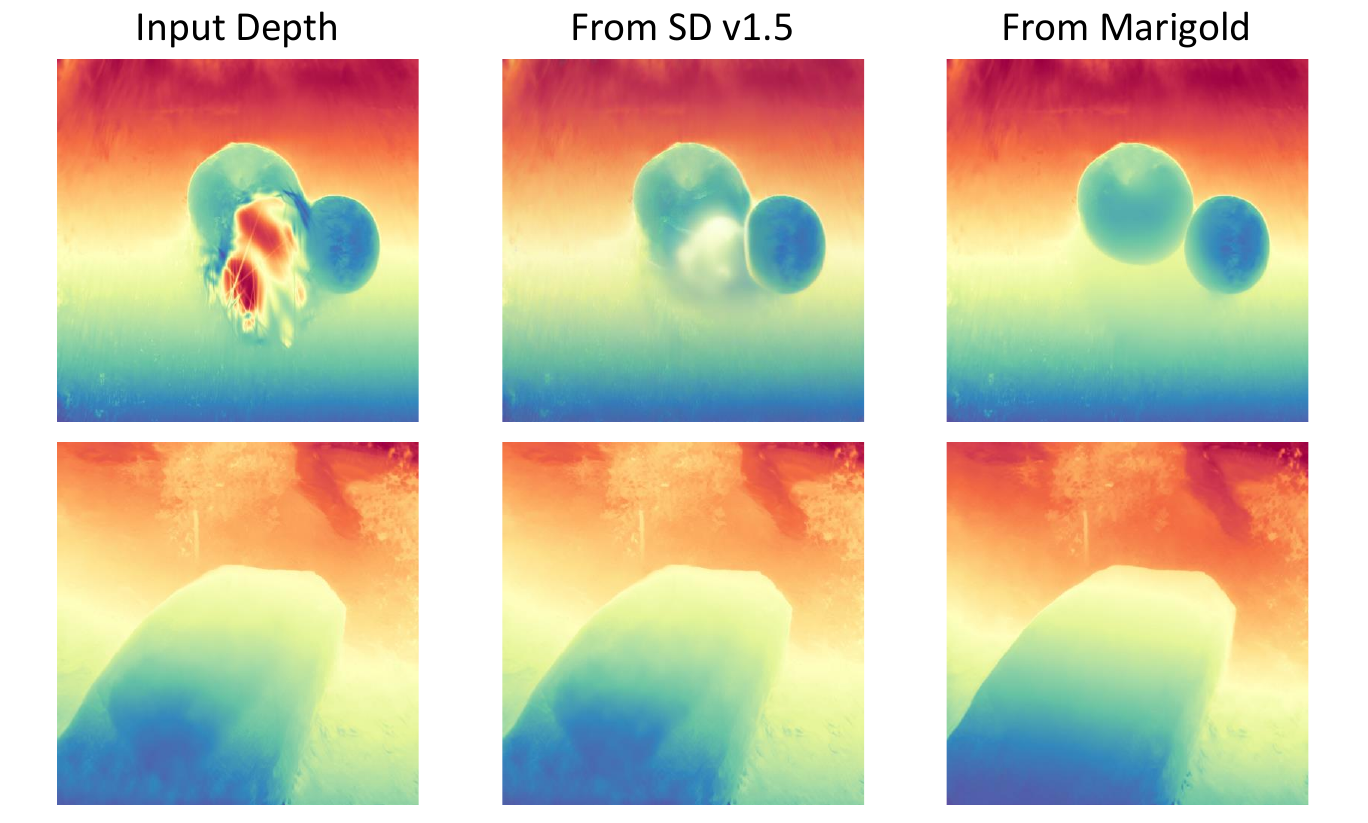}
    \vspace{-9pt}
    \caption{%
        \textbf{Analysis on Pre-trained Weights}. 
    }
    \label{fig:dep}
    \vspace{-20pt}
\end{figure*} 
\subsection{Outliers removal}\label{outlier} 
we unproject depth map and reference image from image space to 3D coordinates to form a colored point cloud. Before this point cloud is merged into original 3D Gaussian point cloud, we need process outliers to improve rendered image quality. To eliminate Gaussian outliers along the edges of the mask, we initially construct a KDTree from the unprojected point cloud. Subsequently, this KDTree is employed to locate the nearest points within the original point cloud, returning points from the original cloud that are within a specified distance threshold.
Subsequently, we utilize the $'remove\_radius\_outlier'$ method from the point cloud data (pcd) library to identify points in the original point cloud that have an insufficient number of neighbors within a specified radius. An intersection of these points and the similar points previously determined using a KDTree is performed, thereby efficiently removing Gaussian outliers at the edges of the mask. Additionally, there are various Gaussian segmentation~\cite{ye2023gaussian, dou2024cosseggaussians, hu2024semantic, cen2023segment, lan20232d} techniques that can be employed for outlier removal, taking advantage of the explicit properties of Gaussian models. Nevertheless, these are not the focal point of the present study and will not be deliberated here.
\section{Analysis} 
\vspace{-30pt}
\begin{figure*}[ht]
    \centering
    \includegraphics[width=0.97\linewidth]{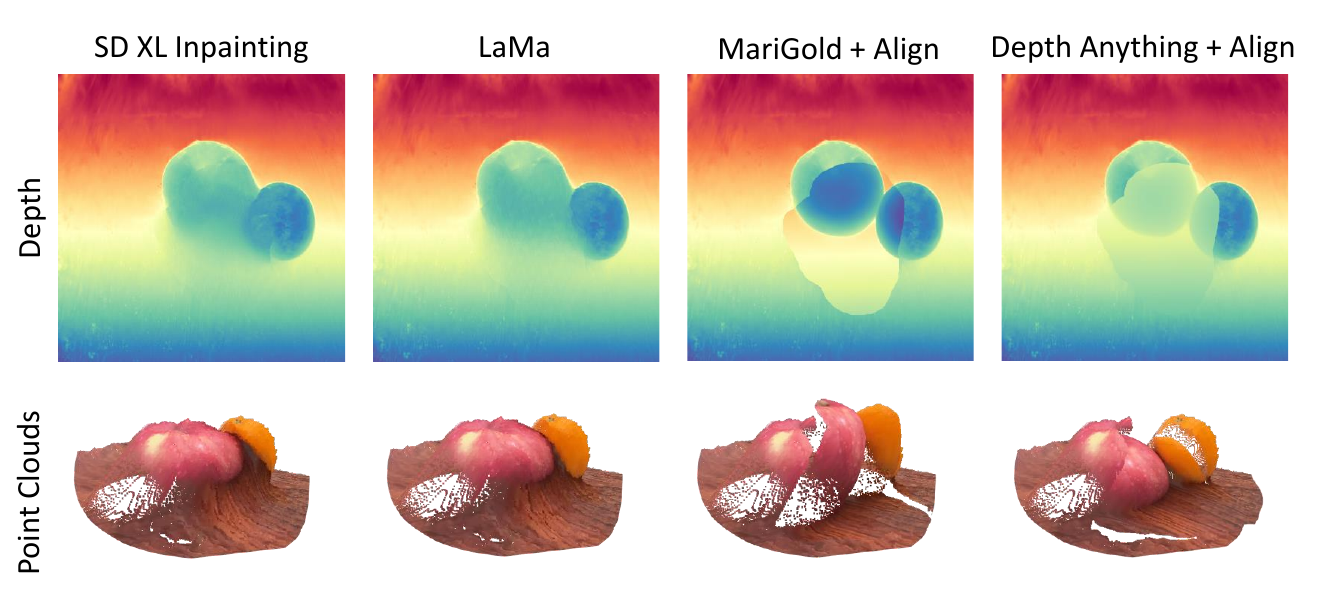}
    % \vspace{-9pt}
    \caption{%
        \textbf{Analysis on Depth Inpainting.} It is evident that the image-based inpainting models, lacking proper guidance, fail to adequately complete the geometric details. Regarding the monocular estimation methods, while a depth alignment method is implemented, they often lead to discontinuities within the inpainted regions
    }
    \label{fig:analysis_1}
    \vspace{-7pt}
\end{figure*}
\subsection{Analysis on pre-trained weights}\label{pretrained}
For the task of depth completion, we employed two distinct sets of initial weights: one derived from Marigold and the other based on Stable Diffusion v1.5. As demonstrated in~\cref{fig:dep}, we display the results under conditions of equivalent data volume and identical training epochs. It is discerned that models initialized with weights from Stable Diffusion v1.5 encountered greater challenges in mastering the depth completion task, a difficulty that was particularly pronounced in complex scenes. In contrast, models that began with Marigold weights exhibited superior proficiency in completing depth, due to prior training on depth maps that reduced the gap between the RGB and depth domains. Following the same training regimen, these models demonstrated an enhanced ability for depth completion and achieved better alignment with the input images.
\begin{figure*}[htp]
    \centering
    \includegraphics[width=0.97\linewidth]{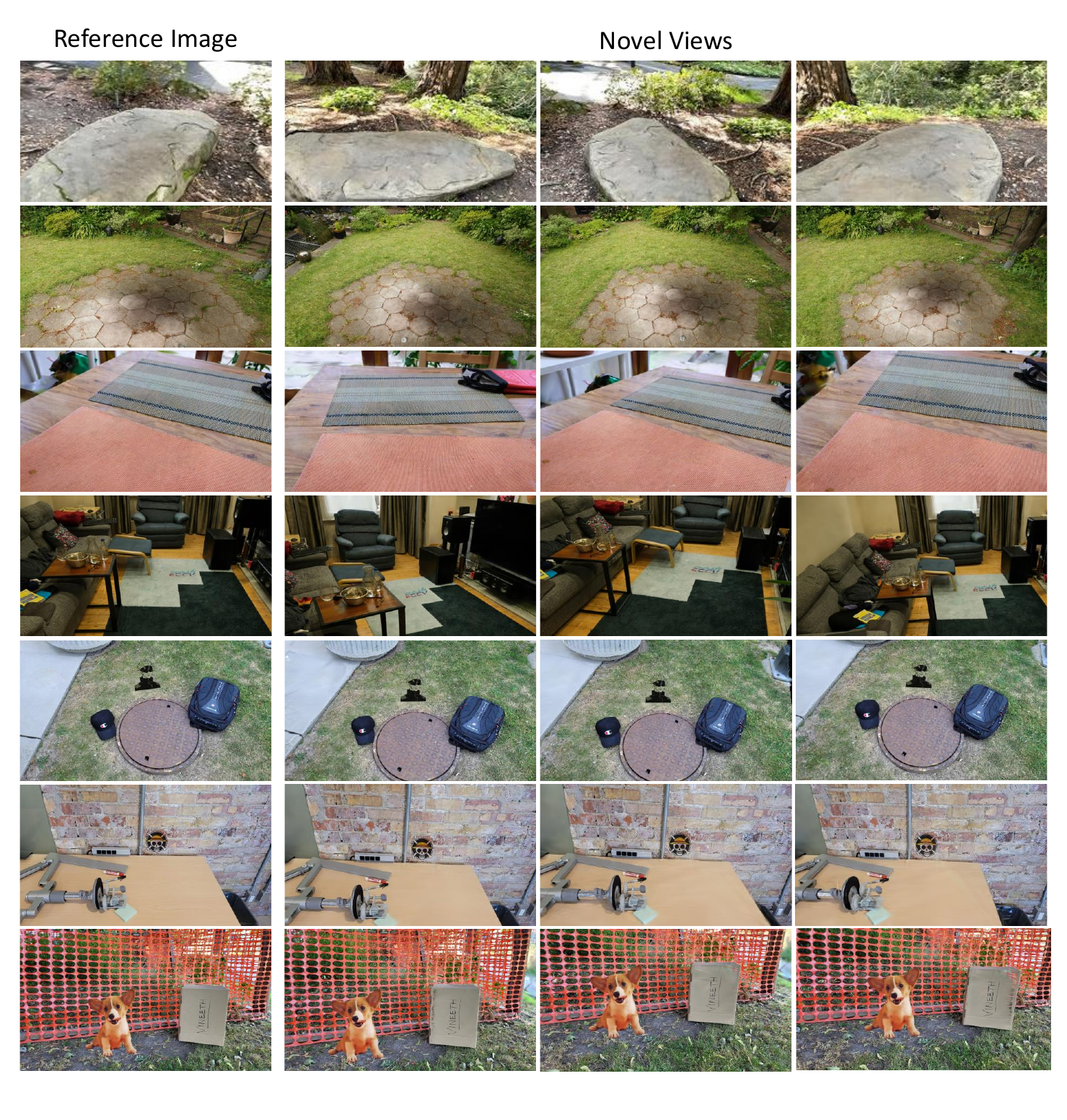}
    % \vspace{-9pt}
    \caption{%
        \textbf{Qualitative Results}. Zoom in for details.  Our method exhibits sharp textures that maintain 3D coherence. We respectfully invite you to view the video featured on the webpage within our supplementary materials
    }
    \label{fig:qualitative_supp}
    % \vspace{-7pt}
\end{figure*} 
\subsection{Analysis on depth inpainting}\label{depth_inpainting]}
We include feature additional results, comparing our method with various cutting-edge baselines, such as SD XL inpainting~\cite{podell2023sdxl} and DepthAnything~\cite{yang2024depth}, with a focus on alignment accuracy. As shown in~\cref{fig:analysis_1}, while SD XL inpainting yields visually appealing results in the RGB domain, a closer inspection of the reprojected point clouds reveals noticeable inaccuracies, akin to those observed in LaMa. Similarly, DepthAnything struggles with discontinuities, leading to a pronounced gap between inpainted areas and their adjacent regions, much like the issues seen with MariGold. Consequently, our learned depth inpainting is critical in securing high-fidelity results.

\section{More Results} 
As shown in~\cref{fig:qualitative_supp}, we present the single reference images for several scenes , along with multiple novel views, to validate the robust 3D consistency achieved by \method. Additionally, we have consolidated all scenes into a webpage included in our supplementary materials and extend an invitation for you to view them.

% \bibliographystyle{splncs04}
% \bibliography{ref.bib}
% \end{document}
% \subsection{More views}